\documentclass[Journal,11pt,NoLineNumbers]{ascelike-new}
\usepackage[utf8]{inputenc}
\usepackage[T1]{fontenc}
\usepackage{lmodern}
\usepackage{graphicx}
\usepackage[figurename=Fig.,labelfont=bf,labelsep=period]{caption}
\usepackage{subcaption}
\usepackage{amsmath}
\usepackage{newtxtext,newtxmath}
\usepackage[colorlinks=true, citecolor=black, linkcolor=black, hidelinks]{hyperref}
\NameTag{Zhuang et al., \today}
\usepackage{relsize} 
\usepackage{amsmath} 
\usepackage{tabularx,booktabs} 
\usepackage{multirow} 
\usepackage{listings} 
\usepackage{subfiles} 
\usepackage{graphicx} 
\usepackage{comment}
\usepackage{relsize}
\usepackage{setspace}

\newcolumntype{C}{>{\centering\arraybackslash} X } 
\graphicspath{{figures/}} 

\usepackage{xcolor}
\definecolor{codegreen}{rgb}{0,0.6,0}
\definecolor{codegray}{rgb}{0.5,0.5,0.5}
\definecolor{codepurple}{rgb}{0.58,0,0.82}
\definecolor{backcolour}{rgb}{0.95,0.95,0.92}
\lstdefinestyle{mystyle}{
    backgroundcolor=\color{backcolour},   
    commentstyle=\color{codegreen},
    keywordstyle=\color{magenta},
    numberstyle=\tiny\color{codegray},
    stringstyle=\color{codepurple},
    basicstyle={\fontsize{9pt}{10.8pt}\ttfamily},
    breakatwhitespace=false,         
    breaklines=true,                 
    captionpos=b,                    
    keepspaces=true,                 
    numbers=left,                    
    numbersep=5pt,                  
    showspaces=false,                
    showstringspaces=false,
    showtabs=false,                  
    tabsize=4
}
\makeatletter
\AtBeginDocument{%
    \def\doi#1{\url{https://doi.org/#1}}}
\makeatother

\usepackage{authblk}

\usepackage{indentfirst} 

\begin{document}



\title{Physics-informed neural networks (PINNs) for numerical model error approximation and superresolution}

\author[1]{Bozhou Zhuang, Ph.D}
\author[2]{Sashank Rana}
\author[3]{Brandon Jones}
\author[4]{Danny Smyl, Ph.D, P.E., M.ASCE$^*$}

\affil[1]{Postdoctoral Researcher, School of Civil and Environmental Engineering, Georgia Institute of Technology, 790 Atlantic Drive, Atlanta, GA 30332-0355.}
\affil[2]{PhD Student, School of Civil and Environmental Engineering, Georgia Institute of Technology, 790 Atlantic Drive, Atlanta, GA 30332-0355.}
\affil[3]{MS Student, School of Civil and Environmental Engineering, Georgia Institute of Technology, 790 Atlantic Drive, Atlanta, GA 30332-0355.}
\affil[4]{Assistant Professor, School of Civil and Environmental Engineering, Georgia Institute of Technology, 790 Atlantic Drive, Atlanta, GA 30332-0355. 
$^*$Corresponding author email address: \textit{danny.smyl@ce.gatech.edu}}

\maketitle

\begin{abstract}
\noindent Numerical modeling errors are unavoidable in finite element analysis. The presence of model errors inherently reflects both model accuracy and uncertainty. To date there have been few methods for explicitly quantifying errors at points of interest (e.g. at finite element nodes). The lack of explicit model error approximators has been addressed recently with the emergence of machine learning (ML), which closes the loop between numerical model features/solutions and explicit model error approximations. In this paper, we propose physics-informed neural networks (PINNs) for simultaneous numerical model error approximation and superresolution. To test our approach, numerical data was generated using finite element simulations on a two-dimensional elastic plate with a central opening. Four- and eight-node quadrilateral elements were used in the discretization to represent the reduced-order and higher-order models, respectively. It was found that the developed PINNs effectively predict model errors in both \(x\) and \(y\) displacement fields with small differences between predictions and ground truth. Our findings demonstrate that the integration of physics-informed loss functions enables neural networks (NNs) to surpass a purely data-driven approach for approximating model errors.
\end{abstract}

\break

\section{Introduction}
\subsection{Dealing with Numerical Model Errors}
\begin{figure*}[h!]
\centering
\includegraphics[trim={2.10cm 0 2.10cm 0},clip,width=5.0in]{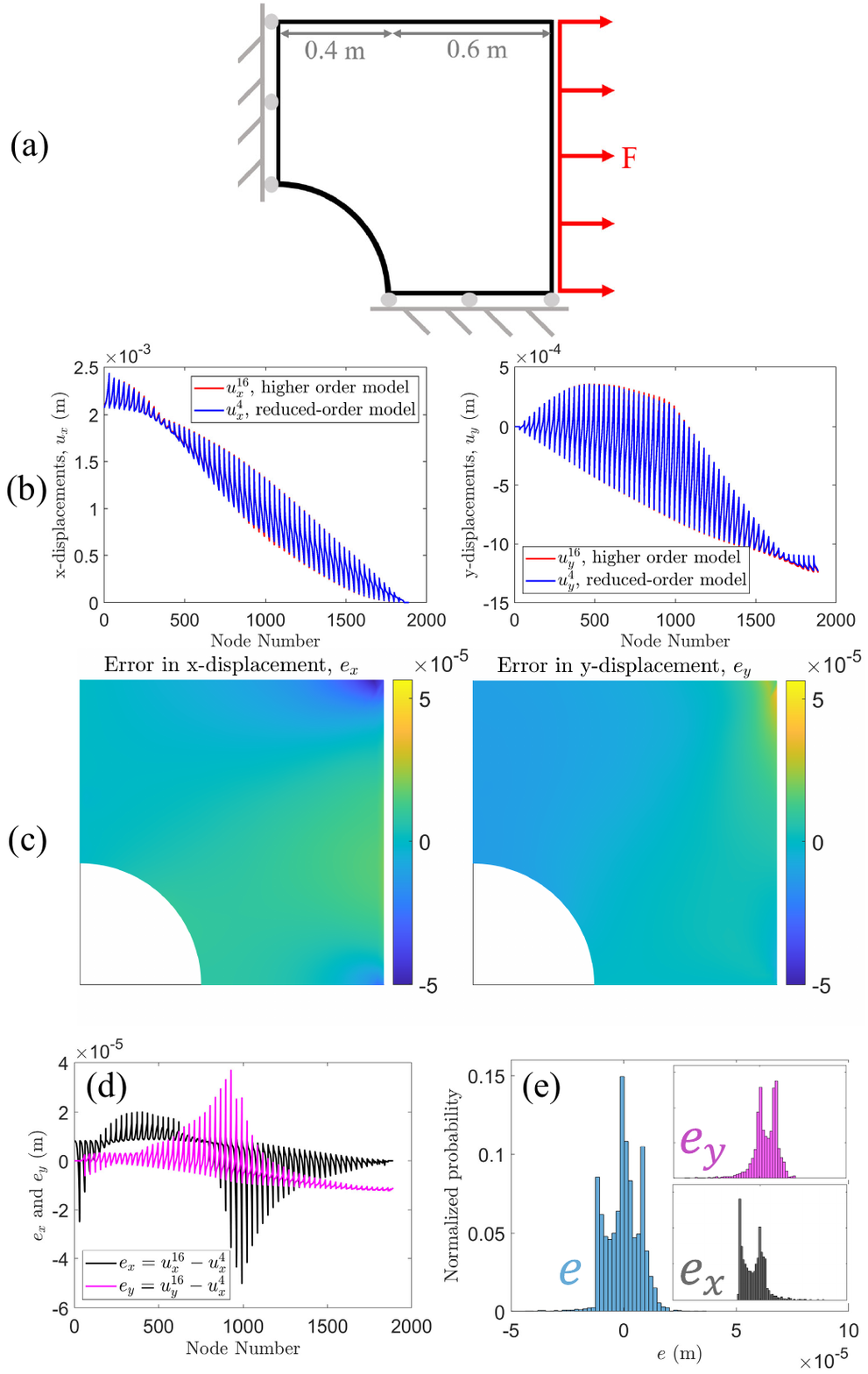}
\caption{Illustration depicting (a) a classic symmetric elastic plate stretching problem solved using 1,800 finite element 4- and 16-node quadrilateral discretizations, denoted with '4' and '16' superscripts, respectively; (b) finite element displacement solutions separated into x and y fields for coincidental nodes; (c) x- and y-displacement error heat maps; (d) the same displacement errors depicted via line graphs; and (e) histograms plotted on the same axis demonstrating typical non-Gaussian model error distributions. Units in (b-e) are in meters. Results in this figure are only for demonstration purposes.}
\label{f1}
\end{figure*}

Numerical model errors are inevitable when using discretized approximations to represent physical domains. \cite{surana2016f}.
Such errors may result from a number of factors, such as errors in approximating a smooth geometry/boundary \cite{reddy2019introduction}, the fidelity of the discretization \cite{babuvska1992h}, the degree of interpolation \cite{duarte1996hp,babuska1981p}, and among others. As a consequence, researchers have aimed to develop tools to measure or approximate numerical model errors. For example, it is common to adopt an absolute error metric that measures the deviation $e$ (i.e., numerical model error) between a solution of a reduced-order numerical model $u_R$ and a higher-order numerical model $u_H$\footnote{We note that analytical solutions are generally not considered as they often do not permit arbitrary geometries or distributions of causal parameters.}:

\begin{equation}
e = u_{H} - u_{R}
\label{error1}
\end{equation}

\noindent where these quantities can be scalars, vectors, or matrices -- herein, the former quantities will be assumed to be vectors henceforth. To demonstrate concepts related to numerical solutions and their errors, Fig. \ref{f1} is provided in the context of the finite element analysis of a two-dimensional plate, which will be introduced later in detail. Here, we can see the differences in the displacement fields generated using higher- and lower-order finite elements models (i.e., Fig. \ref{f1}b) and their respective relative error fields (i.e., Fig. \ref{f1}c). The relatively small errors, compared to displacement magnitudes, are illustrated in the heatmaps and further confirmed by the line graphs in Fig. \ref{f1}d. The non-Gaussian error distributions are shown in Fig. \ref{f1}e.

Commonly, frameworks for dealing with $e$ have included (a) \emph{implicit models}: broadly, the integration of a corrected operator or an operator performing a correction on numerical model quantities of interest and (b) \emph{explicit models}: a model which directly approximates or corrects numerical model solutions, for example, the direct approximation or correction of nodal finite element solutions. This work focuses on solutions to forward problems (cause $\to$ effect), yet explicit and implicit corrections are also applicable in the context of inverse problems (effect $\to$ cause) \cite{arridge2023inverse}.

Classical approaches to error correction have included (a) scalar quantification or error indication for measuring the total error via a norm bound, and (b) probabilistic estimates such as the Bayesian approximation error (BAE) approach \cite{nissinen2007bayesian}. In particular, BAE has been successful in applications with high sensitivity to model errors, especially for medical imaging, and is based on statistical sampling used in integrating mean and covariance quantities into imaging frameworks \cite{kaipio2006statistical}. The BAE approach is often considered as an early approach to data-driven numerical model correction and continues to be used in modern imaging \cite{hauptmann2023model,arridge2023inverse}.

However, there are a number of challenges in implementing classical approaches to deal with numerical model error $e$:

\begin{itemize}
    \item Scalar error quantities do not provide insight on spatial error distributions;
    \item Error indicators may provide only rough bounds and may be biased \cite{freno2019machine};
    \item Statistical approaches may require a prior assumption on the error distribution, which is rarely, if ever, known in the context of numerical model solutions \cite{smyl2021learning}.
\end{itemize}

\noindent In all classical cases mentioned, a key realization is that explicit model errors are not directly quantified and therefore cannot be used to approximate explicit model errors and model compensation.
This poses significant challenges to end users wishing to visualize nodal error distributions, quantify global or local errors when a prior is unknown, and/or quickly assess potential error at a region of interest.

\emph{\textbf{Why haven't we built robust analytical or semi-analytical explicit model error approximators yet?}}
The challenge in developing a model for linking numerical model solutions and/or features (e.g., boundary conditions, spatial distributions of model inputs, source terms, discretization, etc.) to approximations of model errors lies in the complexity of these relations. Not only would such a model be nonlinear, it may be highly ill-conditioned leading to high sensitivity or inaccuracy given small changes to one of many approximator inputs. Thus, it is challenging to construct a sufficiently smooth model error approximator in a well-posedness sense. A well-posed model capable of consolidating a broad set of complex model inputs into a generalizable framework would be highly valuable for numerical model implementation.

It is therefore no surprise that, in at least a general sense, researchers have tested machine learning (ML) models to handle the complex model error approximation task \cite{koponen2021model,smyl2021learning,freno2019machine}.
In doing this, many of the underlying numerical complexities are abstracted to a learning process, which are currently data-driven or black-box implementations. To give context to ML, as it is key to the present work, the following subsection will briefly detail contemporary relevant ML applications.

\subsection{Relevant Applications of ML}
In general, data-driven ML has been successful across the spectrum of science and engineering \cite{brunton2022data}. Research has focused on developing and studying ML approaches for structural design that incorporates a multitude of design variables, response, topology and structural form information to train neural networks (NNs) and later predict structural form \cite{he2021data,shin2020data,shi2020data} and structural topology \cite{hoang2022data,liu2020data,xue2021efficient,lei2019machine,liu2019deep}.
More broadly, structural engineering has benefited from ML in the areas of structural health monitoring (SHM) \cite{farrar2012structural}, performance assessment \cite{sun2020machine} and analysis of various structural causalities/effects \cite{thai2022machine,lee2018background}.

Considering the numerous successes of data-driven ML approaches in structural engineering and mechanics, data-driven only networks often exhibit undesirable properties. It is known that such networks can be sensitive to discrepancies in training data resulting from modeling errors, noise and architecture \cite{gudivada2017data,sessions2006effects}.
The use of regularization and ad hoc techniques, such as dropout \cite{srivastava2014dropout}, offers a means of alleviating the noted sensitivities, since they incorporate limited prior knowledge into the network. As a result, networks may suffer from a lack of generalizability and result in physically unrealistic predictions.

Quite recently, the adoption of a graph neural network (GNN)-based approach for handling physical modeling has been promising. GNNs capture complex relationships and dependencies within data by modeling them as interconnected nodes and edges. This framework allows for a dynamic and flexible representation of both spatial and temporal dependencies. Unlike many NN approaches, GNNs can adapt to evolving structures in the data, facilitating continuous learning and adjustment to changing conditions \cite{zhou2020graph}.

Physics-informed neural networks (PINNs) were previously introduced by Raissi et al. \cite{raissi2019physics} for the purpose of solving physics-governed partial differential equations (PDEs). Designed within supervised ML, PINNs are capable of incorporating specific physics-based prior knowledge into learned models by penalizing unrealistic physical attributes. This is accomplished by applying additional constraints into the learning problem, thereby ensuring that the physics internal to the network are compatible with an accurate physical model. By incorporating simultaneous physics-based prior information and constraints, PINNs have shown promise in enhancing generalizability compared to purely data-driven NNs for supervised emulation of PDE solutions to physical processes. \cite{cai2022physics,karniadakis2021physics,mishra2021estimates}.
These prior works, and the potentiality for using PINNs in solving inverse problems provides the central motivation for this work.

\subsection{Paper Structure and Contributions}
This paper is structured as follows with a focus on modeling errors resulting from discretized civil and mechanical problems.
First, we define the formulation of model error approximation and input for the explicit model error approximator using PINNs.
Next, the details of the numerical simulation and dataset generation are introduced.
Following, the architecture of the developed PINNs and loss functions are presented and discussed.
Then, analysis and discussion of the results will be presented.
Conclusions and limitations will be addressed in the end.

Our contributions in this paper are:

\begin{enumerate}
    \item Propose and test explicit PINN model error approximators;
    \item Provide an in-depth understanding on the sensitivities of explicit machine learned model error approximators considering superresolution, prediction uncertainty, and different loss functions;
    \item Provide pathways forward for future research in ML for explicit numerical modeling errors.
\end{enumerate}

\section{Formulation of Model Error Approximation}

At a practical level, our aims are to develop learned (forward) model error approximators capable of (a) accurately representing distributions of explicit model errors for discretized numerical models and (b) correcting or compensating reduced-order models.
For (b), the intent herein is to increase the approximation accuracy of finite element nodal solutions, in essence "up-scaling" reduced-order nodal solutions to higher-order ones. The specific goal of an explicit machine-learned model error approximator, represented as the function \(\mathcal{A}\), is to approximate errors as a function of selected model inputs $\Theta$:

\begin{equation}
  \mathcal{A}(\Theta) \approx e
\label{NNAE}
\end{equation}

\noindent where we assume that $e$ represents the relative errors described in Eq. \ref{error1}.
In doing this, we assume that the reduced-order model can be corrected or compensated to a more accurate model $\widetilde{u}_{H}$ using:

\begin{equation}
\widetilde{u}_{H} \approx  u_{R} + \mathcal{A}(\Theta)
\label{NNAE2}
\end{equation}

\noindent where we ultimately aim for the following to hold:

\begin{equation}
\widetilde{u}_{H} \approx  u_{H}.
\label{NNAE3}
\end{equation}

\noindent Of course, the precision of $\widetilde{u}_{H}$ will only approach $u_{A}$ as $\mathcal{A}(\Theta)$ remains an approximation.
The error discrepancy resulting from this underscores the fidelity of aims (a) and (b) noted earlier in this subsection. In this study, the outputs from the reduced-order model are used as the input to the PINN for model error approximation, specifically \(\Theta = u_{R}\) \cite{smyl2021learning}.

\begin{figure*}[h!]
\centering
\includegraphics[trim={2cm 3cm 1cm 3cm},clip,width=6.8in]{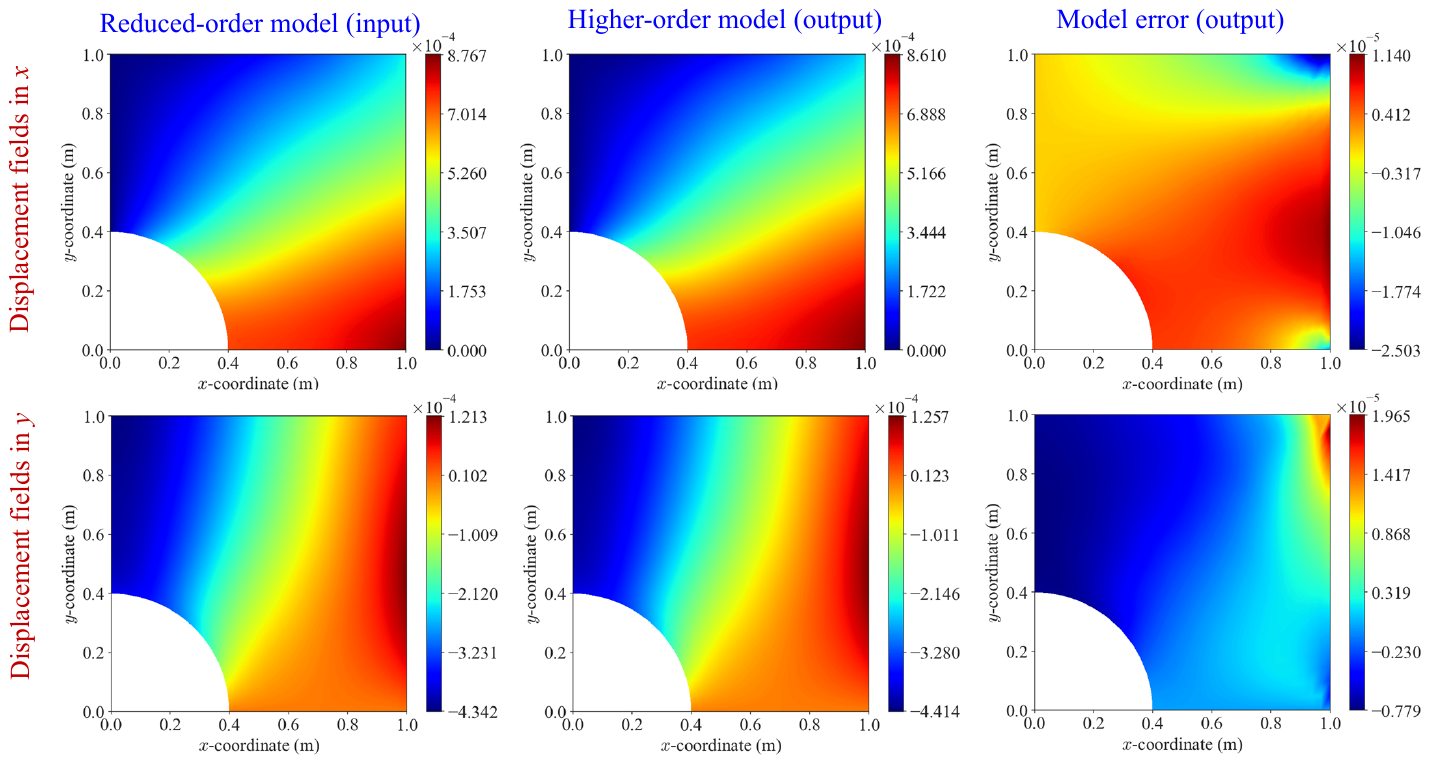}
\vspace{-40pt}
\caption{Displacement fields of lower-order and higher-order models and model error in the generated dataset.}
\label{f2}
\end{figure*}
\section{Numerical Simulation and Dataset Generation}

Finite element analysis was conducted using using an adaptation of \cite{krysl2010thermal}. 
As shown in Fig. \ref{f1}a, a two-dimensional elastic plate with an opening in the center was modeled. The dimension of the plate is $2 \, \text{m} \times 2 \, \text{m}\times 0.005 \, \text{m}$ . The diameter of the opening is 0.8 m. The plate is subjected to a uniformly distributed force along its left and right edges. Due to the symmetric shape and loading conditions, only one-quarter of the plate was considered and symmetric boundary conditions were applied on the symmetrical axes. The forces are randomized in the range of 1 N to 500,000 N per unit length for each generated sample. The uniform elastic modulus is set to \( E_0 = 200\, \text{ GPa} \). To introduce material variability, the elastic modulus were randomized by generating Gaussian random fields with a mean of \( E_0 \) and a standard deviation in the range of \( 0.1E_0 \) and \( 0.5E_0 \). This range of elastic modulus broadly covers metals such as steel and nickel alloys. The correlation lengths in both the \( x \) and \( y \) directions are set to 0.25 m. The Poisson's ratio is 0.28.

The plate was discretized using coarse and fine meshes to represent the reduced-order and higher-order models, respectively. The coarse mesh employs four-node quadrilateral elements (i.e., Q4), and the fine mesh utilizes eight-node quadrilateral elements (i.e., Q8). The reduced-order and higher-order modes have 861 nodes and 9,841 nodes, respectively. The outputs from the from the numerical simulation are the nodal displacement fields from the reduced-order and higher-order models and the model errors in \( x \) and \( y \) displacements. A total of 10,000 samples were generated by randomizing the elastic modulus fields and applied forces. 

One sample from the dataset is visualized in Fig. \ref{f2}. The first, second, and third columns show the displacement fields of the reduced-order models, higher-order models and the model errors, respectively. The first and second rows are the displacement fields in the \( x \) and \( y \) directions, respectively. The displacement in the \( x \)-direction is larger at the bottom edge due to the lower stiffness caused by the circular opening. The right edge shows positive displacement and the left edge has negative displacement  in the \( y\)-direction. The model errors are roughly one order of magnitude smaller than the displacement of the reduced-order or higher-order models. The objectives of this case study are to predict the model errors at the Q4 mesh, shown in the third column, and to predict the displacement fields at the Q8 mesh for superresolution, shown in the second column, using the displacement fields from the reduced-order model as inputs. 

\begin{figure*}[h!]
\centering
\includegraphics[trim={1.5cm 3cm 1.5cm 3cm},clip,width=6.5in]{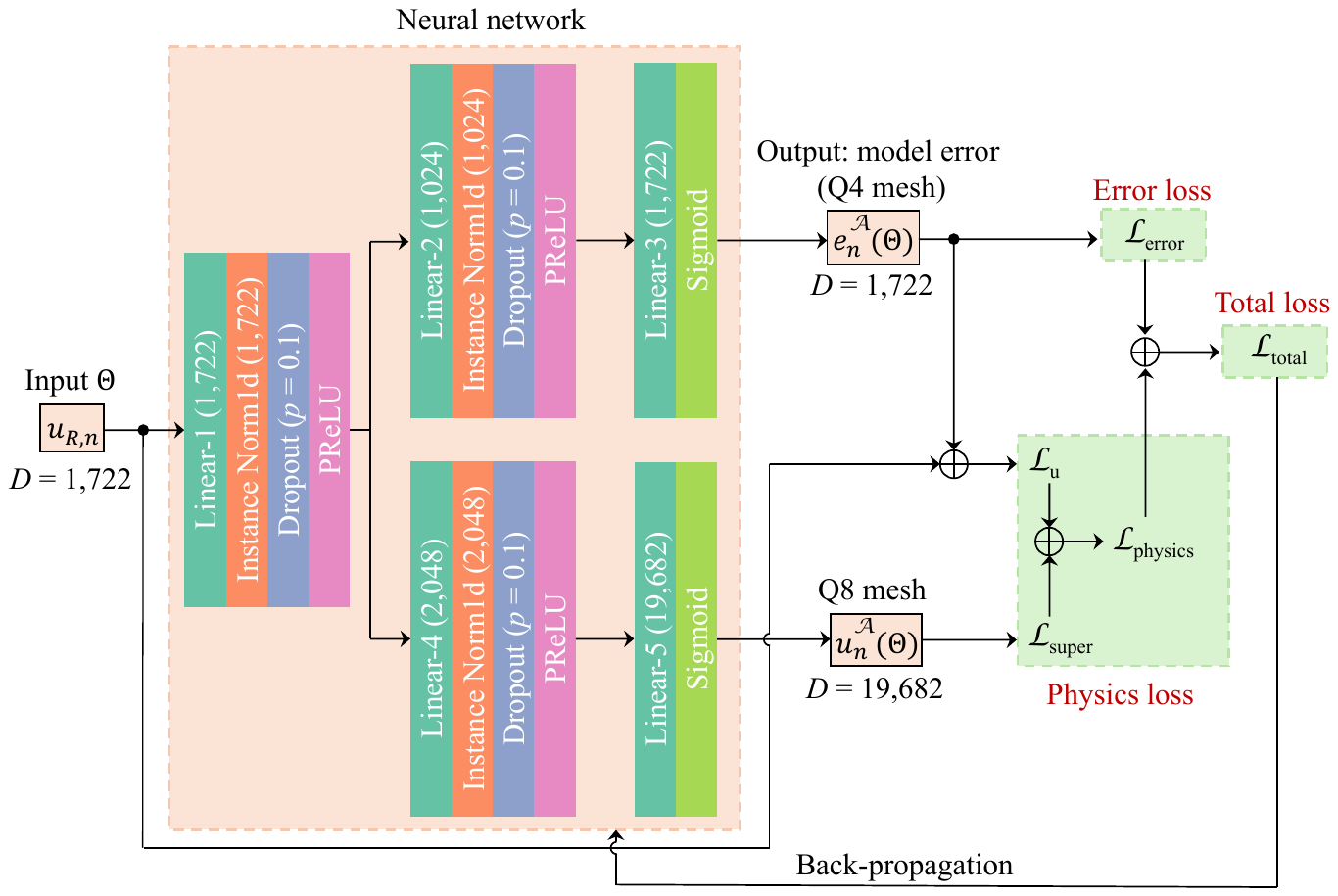}
\vspace{-20pt}
\caption{Architecture of the developed PINN.}
\label{f3}
\end{figure*}
\section{Neural Network Architecture and Training}

\subsection{Network Architecture}
 
The architecture of the neural network integrated here aims to (a) provide reliable predictions of the model errors on the testing data, and (b) maintain stability during training, particularly given the small magnitudes of the model errors considered. To achieve this, we aim for well-generalized networks that produce smooth predictions, thereby ensuring a well-posedness in the problem formulation. 

The architecture of the developed PINN is illustrated in Fig.~\ref{f3}. The input comprises the displacement fields sampled at the nodes of the reduced-order model in the \(x\) and \(y\) directions, as shown in the first column of Fig.~\ref{f2}. Consequently, the input dimension \(D\) is 1,722. The input is processed by a hidden layer (i.e., Linear-1) of the same size, which is then normalized using instance normalization and subjected to a dropout rate of 0.1. The activation function is Parametric ReLU (PReLU), where the negative slopes are learnable parameters. The architecture then diverges into two branches. One predicts the model error with an output dimension of 1,722, and the other predicts the displacement field for a higher-order model using Q8 elements for superresolution of the displacement field with an output dimension of 19,682. Both branches have instance normalization, dropout layers, and PReLU after the hidden layers (i.e., Linear-2 and Linear-4). After the output layers, sigmoid functions with customized upper and lower bounds are applied. The bounds for the first and second branches are \(1.0 \times 10^{-4}\) and \(1.0 \times 10^{-2}\), respectively. 

\subsection{Loss Functions and Model Training}
We formulate the training process by minimizing the total loss \(\mathcal{L}_{\text{total}}\), which incorporates an error loss term \(\mathcal{L}_{\text{error}}\) and a physics-informed loss term \(\mathcal{L}_{\text{physics}}\).  \(\mathcal{L}_{\text{total}}\) and \(\mathcal{L}_{\text{error}}\) can be expressed as

\begin{equation}
\mathcal{L}_{\text{total}} = \mathcal{L}_{\text{error}} + \mathcal{L}_{\text{physics}}
\label{NNr}
\end{equation}

\begin{equation}
\mathcal{L}_{\text{error}} = \frac{1}{N}\sum_{n=1}^N ||e^\mathcal{A}_n(\Theta) - e^d_n||_1
\label{NNr}
\end{equation}

\noindent where \(N\) represents the number of samples; \(e^\mathcal{A}_n\) corresponds to the predicted modeling error from the network given the model parameters \(\Theta\) at sample \(n\); and \(e^d_n\) denotes the true modeling error. \(L_1\) loss is used to quantify the difference between the predictions and ground truth. 
The norm selection was chosen to reduce sensitivity to potential outliers.

Based on preliminary investigation, we identified two physics-informed loss functions that effectively impose additional physical constraints on the model error prediction addressed in this work. To ensure appropriate scaling of each loss term with respect to the error loss, we introduce learnable coefficients \(\beta_1\) and \(\beta_2\) for individual loss components.

\textbf{Displacement Loss Function} \(\mathcal{L_{\text{u}}}\). The displacement loss term is designed to enforce the predicted higher-order displacement field to approximate the actual higher-order displacement field. The loss \(\mathcal{L_{\text{u}}}\) is formulated as: 

\begin{equation}
\mathcal{L_{\text{u}}} = \frac{\beta_1}{N}\sum_{n=1}^{N}  \Vert (e^\mathcal{A}_n(\Theta) + u_{R,n})-u_{(H,\Omega_{Q4}),n}  \Vert _1
\end{equation}

\noindent where \(e^\mathcal{A}_n(\Theta)\) is the predicted error from the network at sample \(n\); \(u_{R,n}\) denotes the displacement field of the reduced-order model at sample \(n\) and is also the input of the network. The summation term (\(e^\mathcal{A}_n(\Theta)\)+\(u_{R,n}\)) represents the predicted displacement field of the higher-order model sampled at Q4 nodes, and \(u_{(H,\Omega_{Q4}),n} \) is the ground truth of the higher-order displacement field sampled on the Q4 mesh (i.e., $\Omega_{Q4}$). It should be mentioned that \(\mathcal{L_{\text{u}}}\) was calculated in the reduced-order dimensionality (i.e., $\Omega_{Q4}$), and the displacement field of higher-order model was interpolated to have the same dimensionality with the reduced-order model. The displacement loss term mitigates potential internal discontinuities in the displacement field, and thus ensures that the PINN remains physically realistic within the defined domain.

\textbf{Superresolution Loss Function \(\mathcal{L_{\text{super}}}\).} The superresolution loss function \(\mathcal{L_{\text{super}}}\) is considered as the second physics-based loss function to directly predict the displacement field of the higher-order model on the Q8 mesh (i.e., $\Omega_{Q8}$). \(\mathcal{L_{\text{super}}}\) is be expressed as

\begin{equation}
\mathcal{L_{\text{super}}}\ = \frac{\beta_2}{N}\sum_{n=1}^{N}   \Vert u^\mathcal{A}_n(\Theta)  - u_{(H,\Omega_{Q8}),n}  \Vert _1
\end{equation}

\noindent where \( u^\mathcal{A}_n(\Theta) \) is the predicted displacement of the Q8 discretization at sample \( n \); \( u_{(H,\Omega_{Q8}),n} \) denotes the true displacement of the Q8 discretization at the sample \( n \).

Gaussian noise with a standard deviation of 0.01\% was added to the training input for regularization purposes. 1,000 samples were reserved for testing and the remaining 9,000 samples were used for training. At each epoch, 4,096 samples were randomly sub-sampled from the training set to improve generalization and manage data throughput effectively. A batch size of 32 was employed. An initial learning rate of \(1.0 \times 10^{-5}\) was used with an exponential decay factor of \(\gamma = 0.99\). Training was conducted using PyTorch on a desktop computer with Nvidia GPU of RTX 4070 up to 300 epochs.
The early stopping was employed to regularize and prevent overfitting of the PINN.

\section{Results and Analysis}

\begin{figure*}[h!]
\centering
\includegraphics[trim={1cm 5cm 1cm 5cm},clip,width=6.5in]{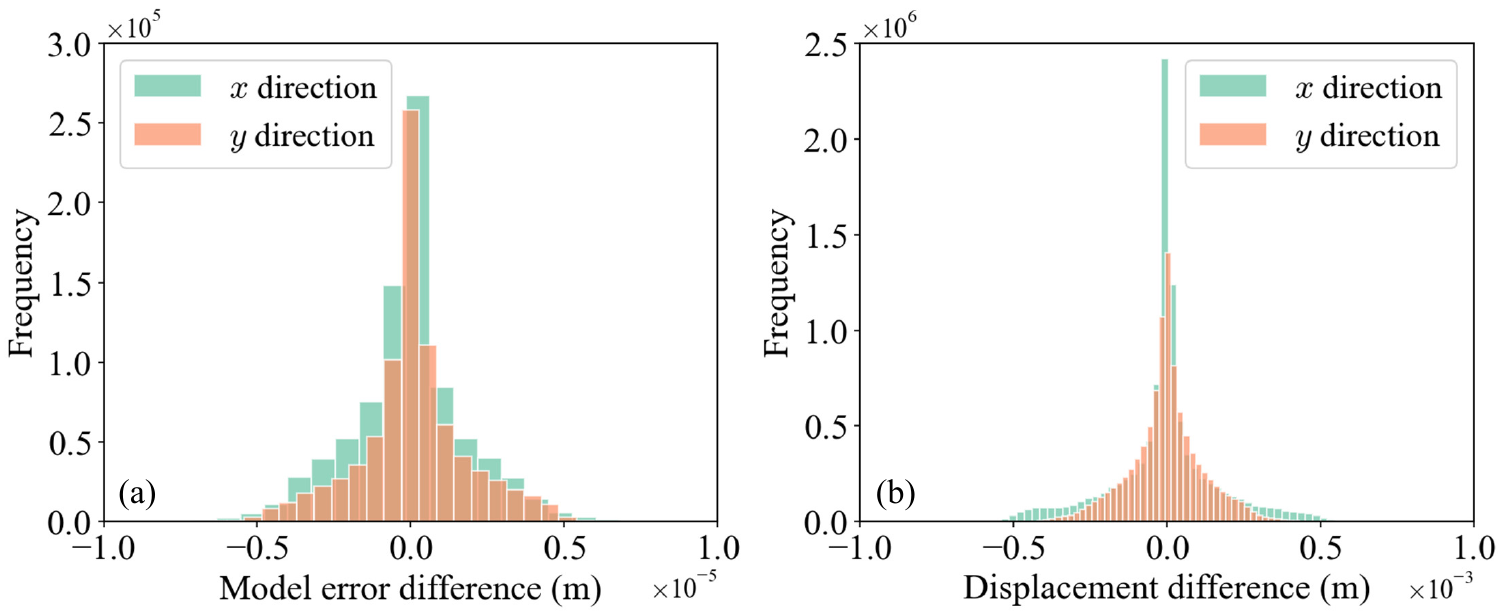}
\vspace{-20pt}
\caption{Distributions of prediction differences of all nodes in the testing set. (a) Mode error difference with Q4 mesh. This histogram is obtained by subtracting the true model error from the predicted model error at all nodes, and (b) displacement field difference with Q8 mesh. This histogram is obtained by subtracting the true displacement from the predicted displacement at all nodes with Q8 mesh for superresolution.}
\label{f4}
\end{figure*}

\begin{figure*}[h!]
\centering
\includegraphics[trim={1cm 3cm 1cm 4cm},clip,width=7in]{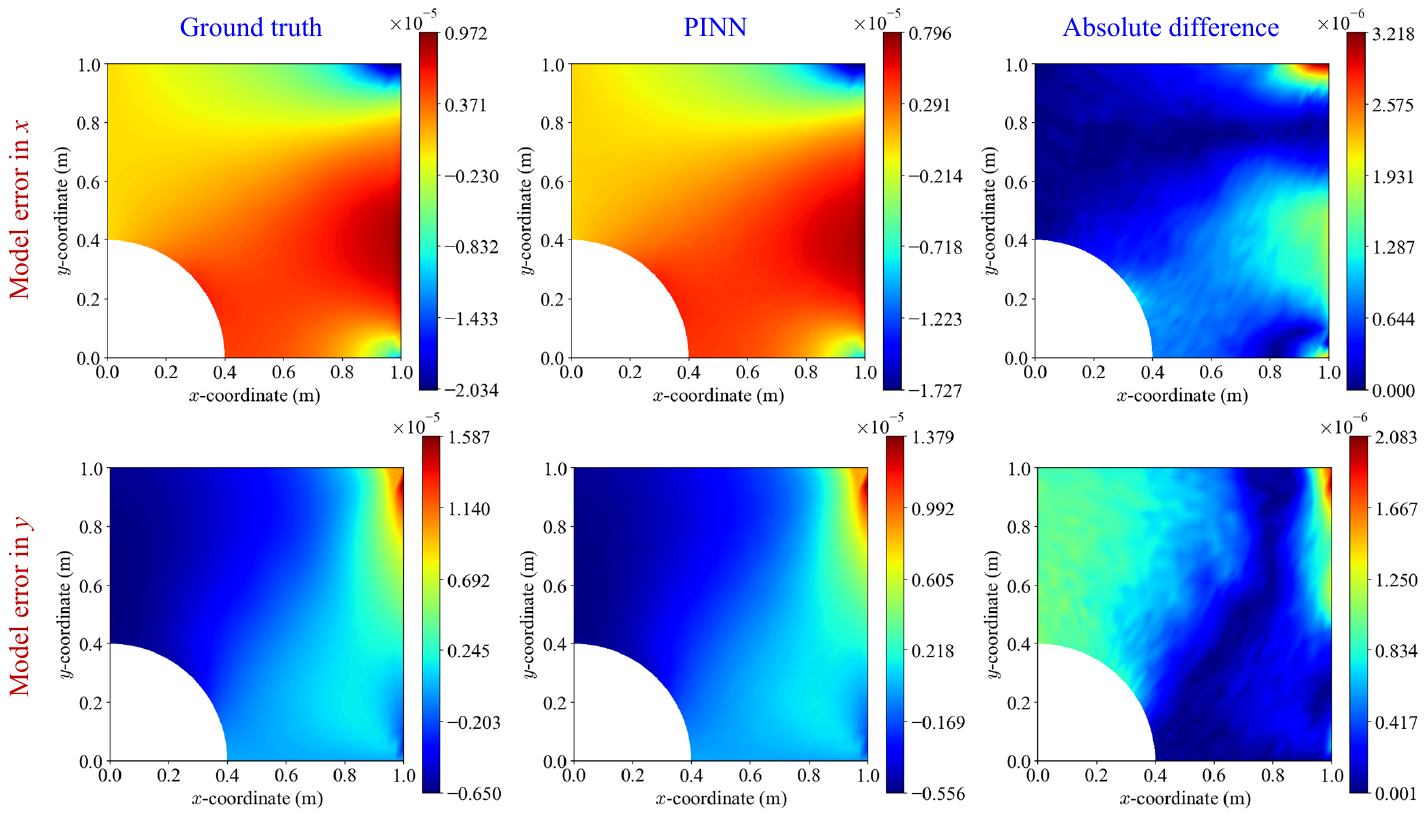}
\vspace{-30pt}
\caption{Comparison of model errors for a single test sample between PINN prediction and ground truth.}
\label{f5}
\end{figure*}

\begin{figure*}[h!]
\centering
\includegraphics[trim={2cm 3.5cm 2cm 3.5cm},clip,width=6.5in]{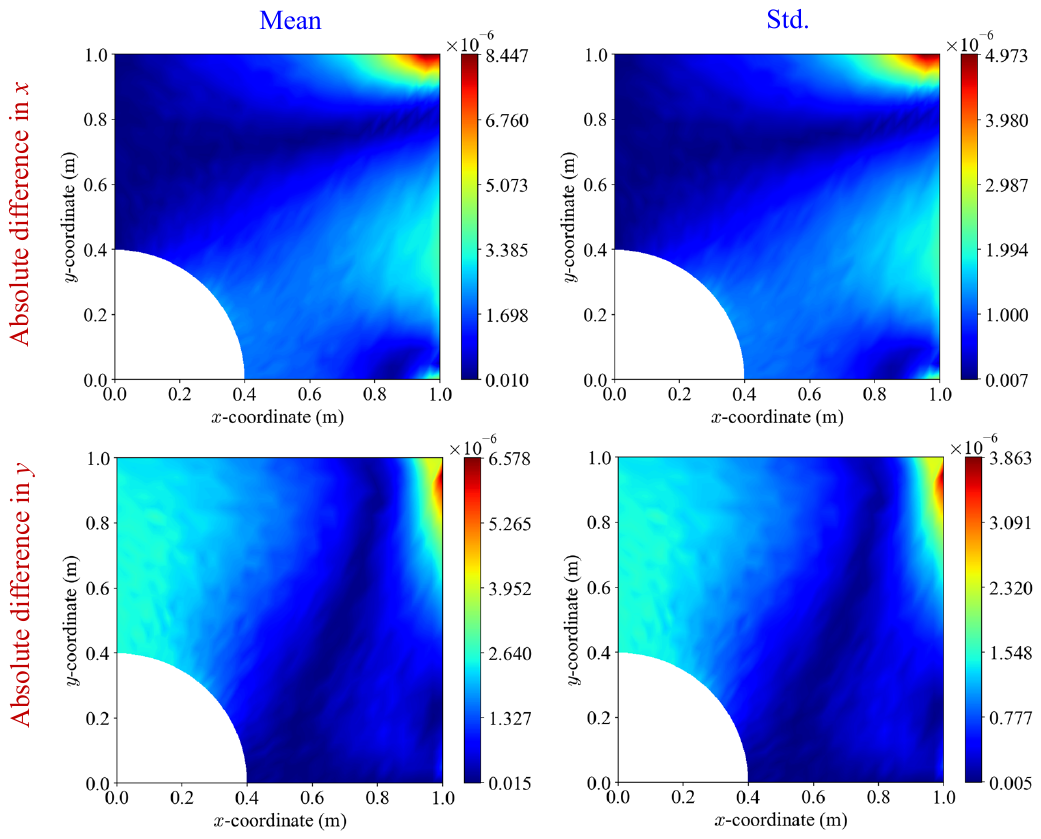}
\vspace{-10pt}
\caption{Absolute difference between predicted model errors and ground truth in the overall testing set.}
\label{f6}
\end{figure*}

\subsection{Model Performance} \label{sec:model_per}
Fig. \ref{f4} shows the prediction difference of model error and Q8 displacement across all nodes in the overall testing set. Specifically, Fig. \ref{f4}a and Figure \ref{f4}b display the histograms of the prediction differences on nodes with Q4 and Q8 meshes, respectively. The
histograms are obtained by subtracting the true model error (or displacement) from the predicted model error (or displacement) at all nodes in the testing set. Therefore, the total bin count in Fig. \ref{f4}a is smaller than that in Fig. \ref{f4}b due to fewer nodes in the Q4 mesh compared to the Q8 mesh. The histograms are symmetrically centered around zero, which indicates that most estimates closely approximate the model error and Q8 displacement to their respective ground truths in both the \(x\) and \(y\) directions.

Fig. \ref{f5} displays the model error prediction for a random sample in the testing set and its ground truth. The first and second columns in Fig. \ref{f5} represent the ground truth and PINN prediction, respectively. The third column is the absolute difference between the PINN prediction and the ground truth at each node. The developed PINN can accurately predict the model error in both the \(x\) and \(y\) directions. The absolute differences are approximately one order of magnitude smaller than the model error at corresponding positions. To visualize the absolute difference across all testing samples, the mean and standard deviation (std.) were taken for all testing samples as shown in Fig. \ref{f6}. The first and second columns of Fig. \ref{f6} show the mean and standard deviation, respectively, and the first and second rows are for \(x\) and \(y\) directions, respectively. Notably, the mean absolute difference is less than \(4.0 \times 10^{-6}\) m in most areas. The PINN shows larger errors when predicting model errors at nodes located in the top right corner. This could be induced by the free boundaries in this region.

\begin{figure*}[h!]
\centering
\includegraphics[trim={2cm 3cm 1cm 3cm},clip,width=7.3in]{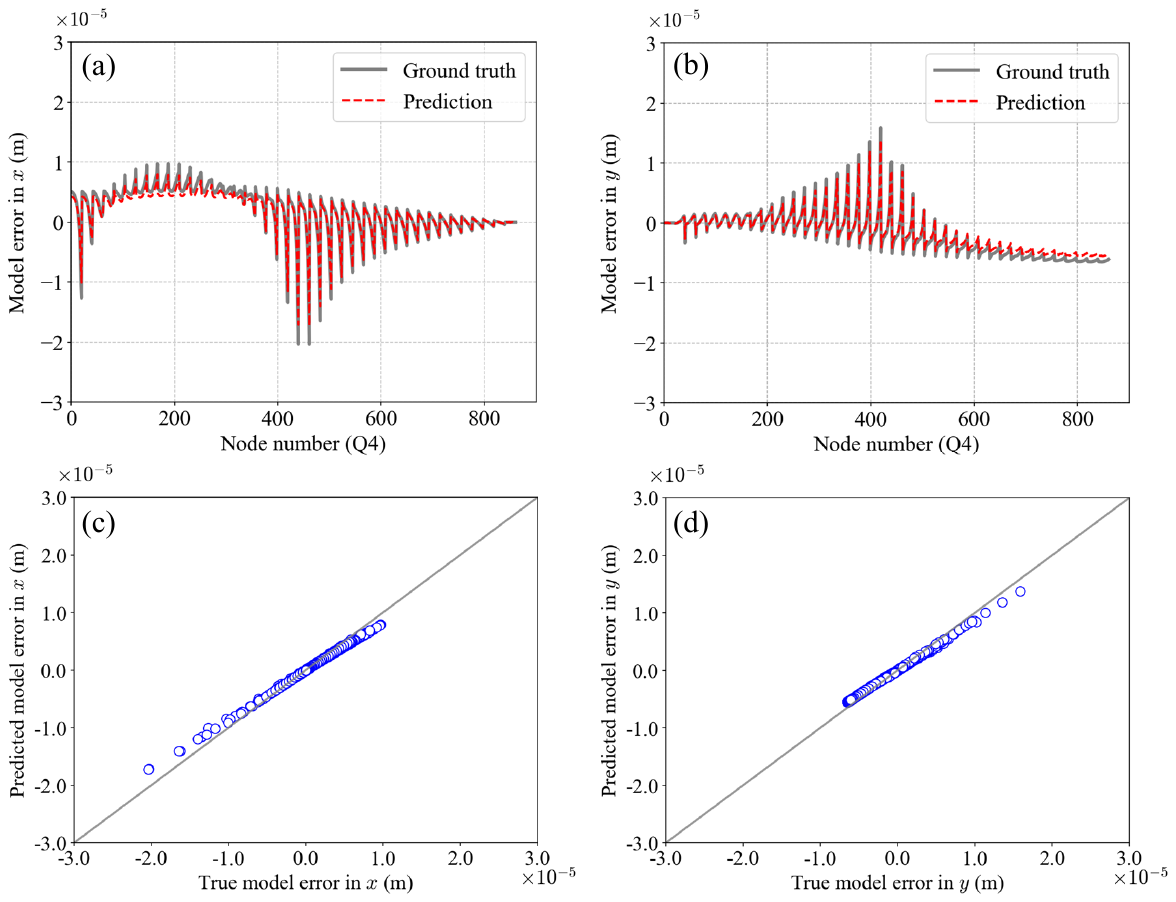}
\vspace{-10pt}
\caption{Nodal comparisons of model errors between PINN prediction and ground truth of the same sample in Fig.~\ref{f5}. (a) and (c): model errors in \(x\), (b) and (d): model errors in \(y\).}
\label{f7}
\end{figure*}

\begin{figure*}[h!]
\centering
\includegraphics[trim={2cm 3cm 1cm 3cm},clip,width=7.3in]{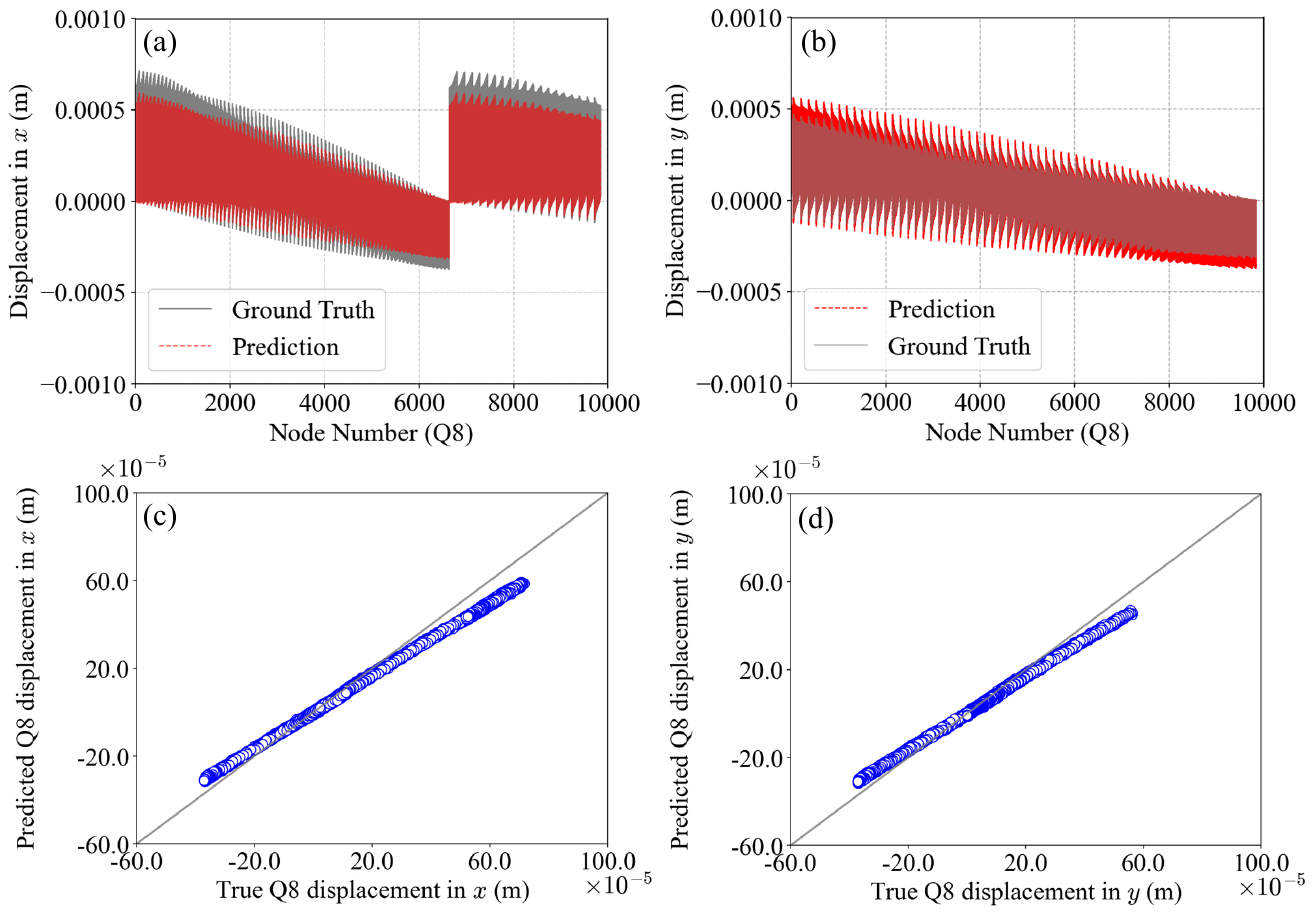}
\vspace{-20pt}
\caption{Nodal comparisons of Q8 displacement between PINN prediction and ground truth of the same sample in Fig. ~\ref{f5}. (a) and (c): displacement in \(x\), (b) and (d): displacement in \(y\).}
\label{f8}
\end{figure*}

Figs. ~\ref{f7} and ~\ref{f8} show quantitative nodal comparisons for model errors and Q8 displacement, respectively. The same sample from Fig. ~\ref{f6} was used for consistency. The model error predictions closely approximate the ground truth. However, as shown in Fig. ~\ref{f8}, the PINN slightly underestimates the Q8 displacement for this specific testing sample.

\subsection{Effects of Physics-informed Loss Functions}

\begin{table*}[h!]
\centering
\caption{Effects of physics-informed loss functions on the model error loss}
\resizebox{\textwidth}{!}{\begin{tabular}{ccccc}
\hline
\text{Cases} & \text{Main Loss Function}  & \text{Physics-informed Loss Function(s)} & \text{Training \(\mathcal{L}_{\text{error}}\) (\(\times 10^{-6}\) m)} & \text{Testing \(\mathcal{L}_{\text{error}}\) (\(\times 10^{-6}\) m)} \\
\hline

1 & \(\mathcal{L}_{\text{error}}\) & \(\mathcal{L}_{\text{u}}\) + \(\mathcal{L}_{\text{super}}\) & \( 2.30 \pm 0.00 \) & \( 1.38 \pm 2.10 \) \\

2 & \(\mathcal{L}_{\text{error}}\) & \(\mathcal{L}_{\text{u}}\) & \( 2.20 \pm 0.00 \) & \( 1.14 \pm 2.90 \) \\

3 & \(\mathcal{L}_{\text{error}}\) & - & \( 2.50 \pm 0.00 \) & \( 1.23 \pm 2.30 \) \\

\hline
\end{tabular}}
\label{t1}
\end{table*}

\begin{figure*}[h!]
\centering
\includegraphics[trim={1cm 7cm 1cm 7cm},clip,width=6.8in]{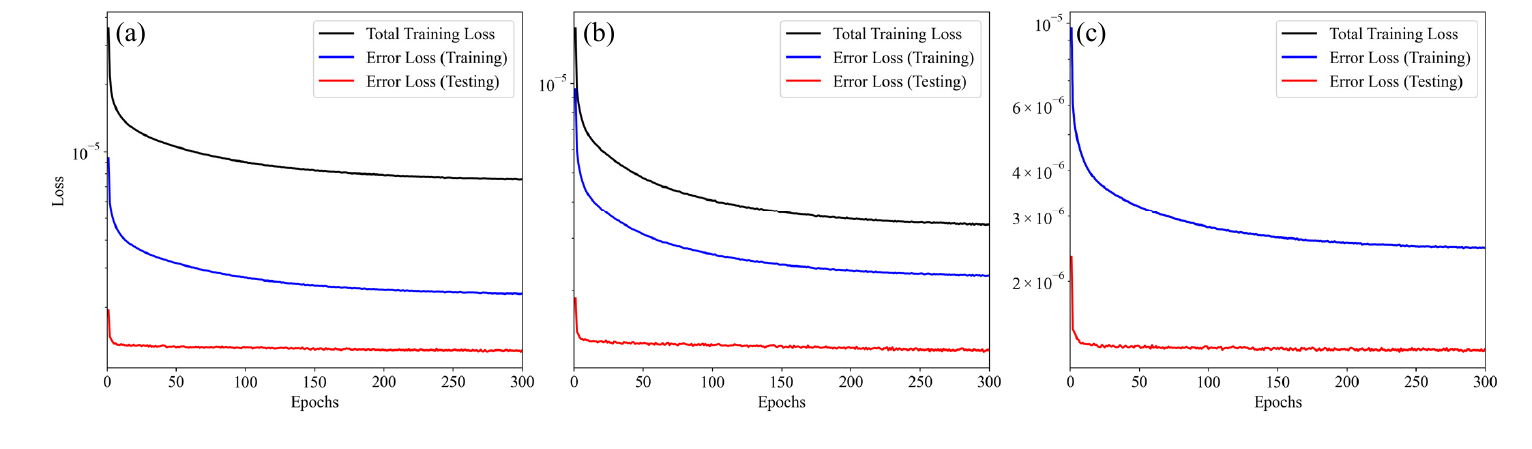}
\vspace{-10pt}
\caption{Loss curves of cases listed in Table \ref{t1}. (a) Case 1, (b) Case 2, and (c) Case 3.}
\label{f9}
\end{figure*}

\begin{figure*}[h!]
\centering
\includegraphics[trim={1cm 3.5cm 1cm 3.5cm},clip,width=7.3in]{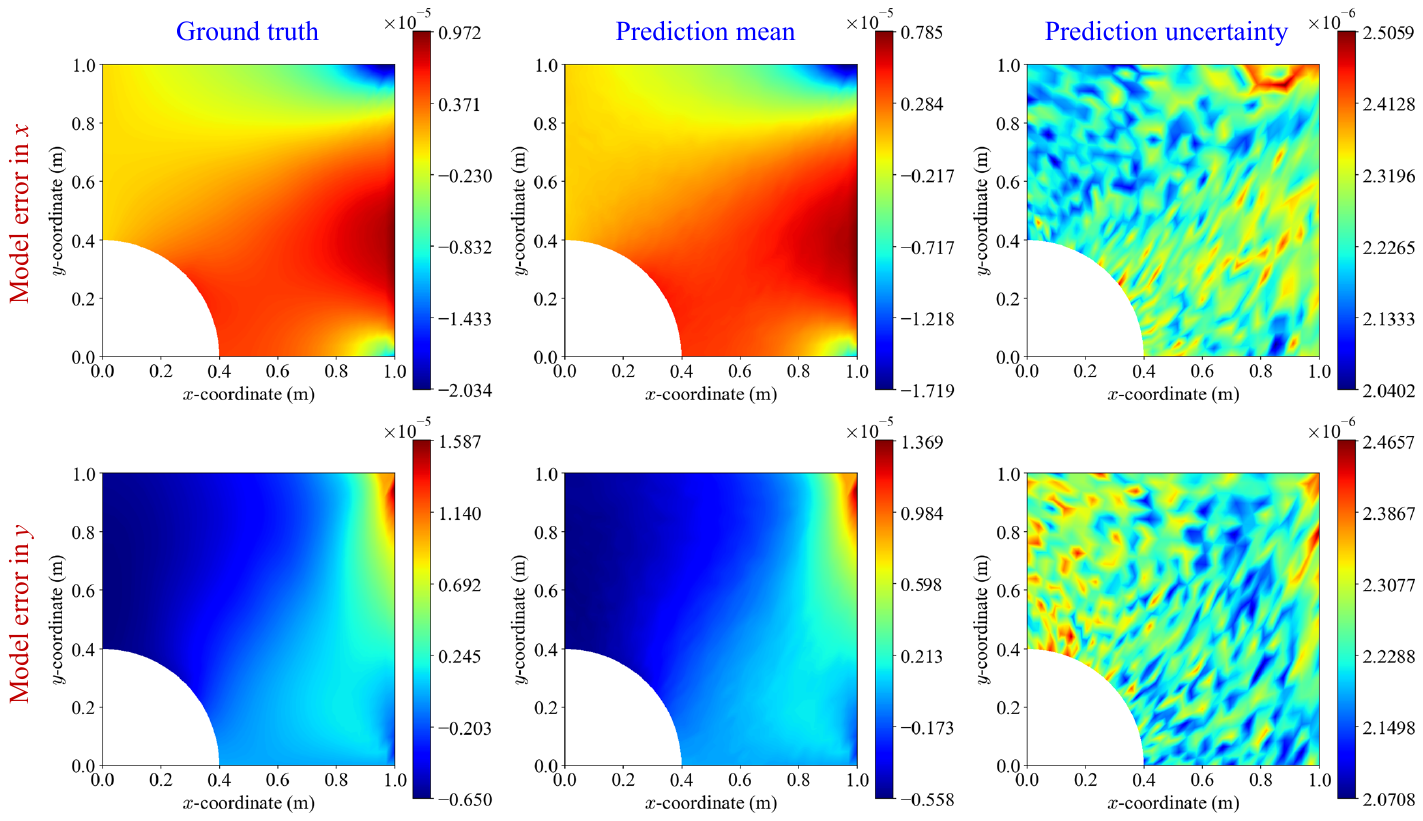}
\vspace{-20pt}
\caption{Prediction uncertainty of the same testing sample in Fig. ~\ref{f5}.}
\label{f10}
\end{figure*}

The physics-informed loss functions were sequentially removed to evaluate their impact on prediction accuracy. Models were trained with the combinations of loss functions listed in Table \ref{t1}. Case 1 represents the baseline PINN discussed in Section~\ref{sec:model_per}. In Case 2, only the displacement loss function \(\mathcal{L}_{\text{u}}\) was retained as the physics-informed loss function and the superresolution loss function \(\mathcal{L}_{\text{super}}\) was removed. For Case 3, \(\mathcal{L}_{\text{u}}\) was also removed and thus there was no physics-informed loss function. The model error loss \(\mathcal{L}_{\text{error}}\) remained as the main loss function across all cases. 

The mean and standard deviation of the training and testing model error loss over the last ten epochs are presented in Table \ref{t1}. The loss curves of three cases are shown in Fig. \ref{f9}. Comparing Case 1 and Case 2, adding the superresolution loss function \(\mathcal{L}_{\text{super}}\) in Case 1 slightly increases both the training and testing losses. The mean training loss rises from \(2.20 \times 10^{-6}\) m in Case 2 to \(2.30 \times 10^{-6}\) m in Case 1, and the mean testing loss increases from \(1.14 \times 10^{-6}\) m to \(1.38 \times 10^{-6}\) m. This occurs because \(\mathcal{L}_{\text{super}}\) does not depend on the model error directly. However, adding the superresolution physics term remains beneficial since it directly outputs the displacement field at the Q8 mesh to provide higher-resolution displacement relative to the Q4 discretization alone. Comparing Case 2 with Case 3, including \(\mathcal{L}_{\text{u}}\) reduces both training and testing losses and thus improves the accuracy. The mean training loss drops by 12.0\% from \(2.50 \times 10^{-6}\) m in Case 3 to \(2.20 \times 10^{-6}\) m in Case 2, while the mean testing loss reduces by 7.3\% from \(1.23 \times 10^{-6}\) m to \(1.14 \times 10^{-6}\) m. This improvement demonstrates that \(\mathcal{L}_{\text{u}}\) poses additional constraints that guide the higher-order displacement predictions to better approximate the true displacement field at Q4 nodes. Therefore, the addition of the displacement error loss \(\mathcal{L}_{\text{u}}\) enhances the network’s capability beyond pure data-driven input-output mapping.

\subsection{Prediction Uncertainty}

 Dropout layers are commonly used to prevent overfitting, and they can also be used to approximate Bayesian inference to assess prediction uncertainty of deep neural networks \cite{gal2016dropout}. Previous work has demonstrated the effectiveness of dropout layers in introducing model uncertainties for deep learning models \cite{zhuang2024impurity,arcaro2024damage}. 
 
 In this subsection, the dropout layers were kept active during testing, and the trained PINN, with both \(\mathcal{L}_{\text{u}}\) and \(\mathcal{L}_{\text{super}}\) as physics-informed loss functions, was applied 2,000 times on the same testing sample to estimate prediction uncertainty. The mean and prediction uncertainty, represented by the standard deviation across the 2,000 runs, are presented in Fig. \ref{f10}. 

The PINN demonstrates a high level of confidence in predicting model error in both the \(x\) and \(y\) directions. The prediction uncertainty is within the range of \(2-2.5 \times 10^{-6}\) m, which is much smaller than the actual model error values in both directions. Additionally, the prediction uncertainty does not show a clear pattern on the plate. This could be attributed to the ideal dataset generation process.

\subsection{Superresolution}

\begin{figure*}[h!]
\centering
\includegraphics[trim={0cm 8cm 0cm 5cm},clip,width=7in]{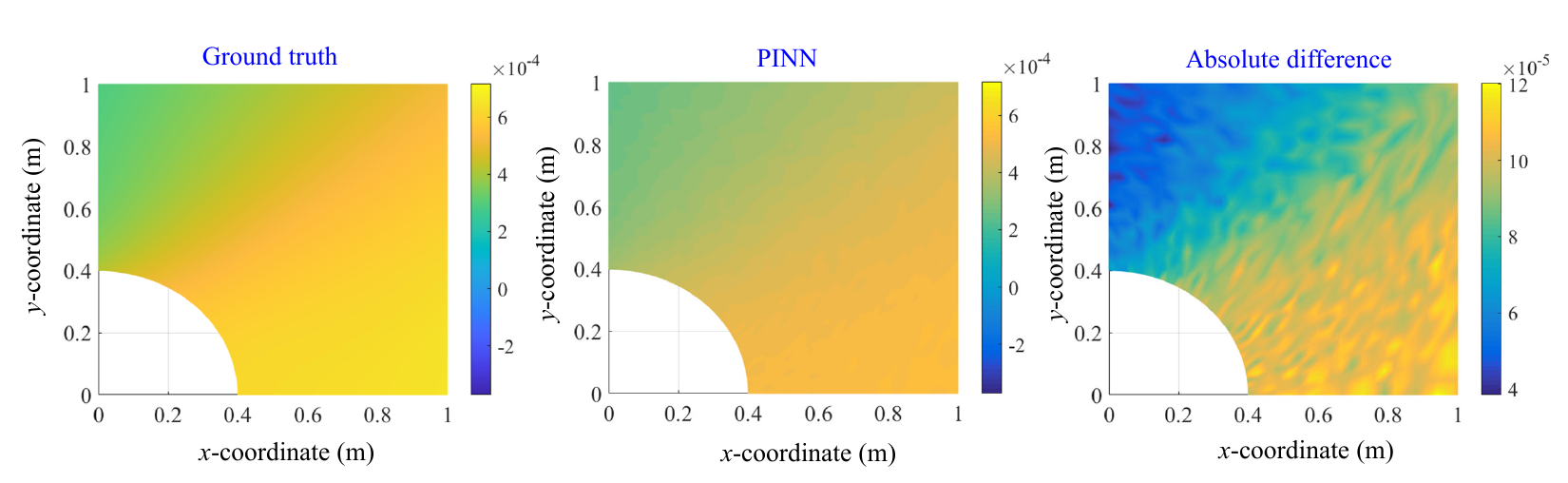}
\vspace{-20pt}
\caption{Superresolution of the \(x\)-displacement field with Q8 discretization using the same testing sample in Fig. ~\ref{f5}.}
\label{f11}
\end{figure*}

A benefit of the proposed PINN approach proposed herein is the superresolution, namely the upsampling of a reduced-order finite element solution to a higher-order solution approximation (in both the $h$ and $p$ sense).
While the fidelity improvement in \( h \) (i.e., fineness of mesh) is analogous to GPU upscaling in video rendering—such as Deep Learning Super Sampling (DLSS)—the concept of super-resolving to higher interpolating polynomial spaces \( p \) using PINNs is novel \cite{babuvska1992h,duarte1996hp,surana2016f}.
As shown in Fig. \ref{f8}, we observe a satisfactory matching of true and predicted solutions while simultaneously attaining information on predicted model errors.
The true and predicted displacement fields of superresolution in the \( x\) direction are visualized in Fig. \ref{f11}.
While these images do provide more spatial information than the compensated reduced-order model $\widetilde{u}_{H} \approx  u_{R} + \mathcal{A}(\Theta)$, additional improvements to the network architecture is needed. More data is likely required and a broader search of the hyperparamater space should be investigated in order to improve superresolved solution accuracy.

\section{Conclusions}
In this paper, PINNs were centrally developed to (a) approximate numerical model errors between reduced-order and higher-order finite element model and (b) provide superresolution approximations of higher-order ($h$ and $p$) solutions. Numerical data was generated using finite element simulations on a two-dimensional elastic plate with a central opening. The key findings are as follows:

\begin{itemize}
    \item The developed PINNs effectively predict model errors in both \(x\) and \(y\) displacement fields with small differences between predictions and ground truth. This highlights the potential for enhancing reduced-order models by approximating errors explicitly with ML.
    \item The integration of physics-informed loss functions enables NNs to go beyond a purely data-driven approach for approximating model errors. Incorporating the superresolution loss function \(\mathcal{L}_{\text{super}}\) enables direct prediction of higher-order displacement fields. The displacement loss function \(\mathcal{L}_{\text{u}}\) improves the model’s accuracy by posing additional constraints that guide the higher-order displacement predictions to better approximate the true displacement field.
    \item Prediction uncertainty analysis with dropout layers indicates that the PINN is confident in its predictions for the model error. The prediction uncertainties in both the \(x\) and \(y\) directions are within the range of \(2-2.5 \times 10^{-6}\) m, which are much smaller than the actual model error. The prediction uncertainty does not show a clear spatial pattern. This could be attributed to the controlled nature of the dataset generation process.
\end{itemize}

This study only focuses on the model error between Q4 and Q8 discretizations with a specific geometry and boundary condition. The variations in meshes, geometries, and boundary conditions, and material properties (aside from the randomized distributions of the elastic modulus) are not considered. For the model to be better generalized, future work could focus on establishing a meta numerical data set with a broader range of meshes, geometries, boundary conditions, and material variability.
From a NN implementation standpoint, the use of attention layers, residual blocks, and a suite of modern techniques may improve PINN performance.
Lastly, more physics-informed loss function should be investigated for improved accuracy.

\section*{Data Availability Statement}
\noindent Some or all data, models, or code that support the findings of this study are available from the corresponding author upon reasonable request.

\section*{Acknowledgment} 

The authors would like to acknowledge the School of Civil and Environmental Engineering at Georgia Institute of Technology for supporting this work.






\begin{spacing}{1.5}
\bibliography{ascexmpl-new}
\end{spacing}

\end{document}